\documentclass{article}
\usepackage{amsmath,epsfig}
\usepackage[preprint]{spconfa4}
\usepackage{xcolor}
\usepackage{cite}
\usepackage{amsmath,epsfig}
\usepackage{graphicx}
\usepackage{amsmath}
\usepackage{amsthm}
\usepackage{booktabs}
\usepackage{algorithm}
\usepackage{algorithmic}
\usepackage{graphicx}
\usepackage{amsmath}
\usepackage{amsmath,amssymb,amsfonts}
\usepackage{algorithm}
\usepackage{booktabs}
\usepackage{algorithmic}
\usepackage{bm}
\usepackage{multirow}
\usepackage{amssymb}
\usepackage{color}
\usepackage{subcaption}
\usepackage{multirow}
\usepackage{amssymb}
\usepackage{subcaption}
\usepackage{mathtools}
\usepackage{makecell}
\usepackage{enumitem}
\usepackage{caption}

\let\OLDthebibliography\thebibliography
\renewcommand\thebibliography[1]{
  \OLDthebibliography{#1}
  \setlength{\parskip}{0pt}
  \setlength{\itemsep}{0pt plus 0.3ex}
}

\begin{document}\sloppy

\def\x{{\mathbf x}}
\def\L{{\cal L}}

\title{Localizing Semantic Patches for Accelerating Image Classification}
%
\name{Chuanguang Yang$^{\dagger \ddagger}$ \qquad Zhulin An$^{\dagger}$\sthanks{Corresponding author.} \qquad Yongjun Xu$^{\dagger}$}
\address{$^{\dagger}$Institute of Computing Technology, Chinese Academy of Sciences, Beijing, China \\
	$^{\ddagger}$University of Chinese Academy of Sciences, Beijing, China \\
	\{yangchuanguang, anzhulin, xyj\}@ict.ac.cn}

\maketitle

\begin{abstract}
	Existing works often focus on reducing the architecture redundancy for accelerating image classification but ignore the spatial redundancy of the input image. This paper proposes an efficient image classification pipeline to solve this problem. We first pinpoint task-aware regions over the input image by a lightweight patch proposal network called AnchorNet. We then feed these localized semantic patches with much smaller spatial redundancy into a general classification network. Unlike the popular design of deep CNN, we aim to carefully design the Receptive Field of AnchorNet without intermediate convolutional paddings. This ensures the exact mapping from a high-level spatial location to the specific input image patch. The contribution of each patch is interpretable. Moreover, AnchorNet is compatible with any downstream architecture. Experimental results on ImageNet show that our method outperforms SOTA dynamic inference methods with fewer inference costs. Our code is available at https://github.com/winycg/AnchorNet.
\end{abstract}
\begin{keywords}
	Patch Localization, Dynamic Acceleration, Image Classification
\end{keywords}
\section{Introduction}
Inference acceleration of a given CNN for solving vision tasks is a practical problem in real-world applications. Most existing works aim to optimize the model complexity directly. Typical methods are mainly divided into efficient network architecture design~\cite{Mingxing2019EfficientNet,yang2020gated}, network pruning~\cite{yang2019multi,cai2022prior} and knowledge distillation~\cite{hinton2015distilling,yang2021hierarchical,yang2022mutual}. The above works can be summarized as static model compression methods, which use the same inference path for all input samples. However, various input samples often have different difficulties for classification~\cite{huang2017multi}. Another vein lies in the dynamic inference scheme~\cite{huang2017multi,teja2018hydranets} that adaptively adjusts the network architecture to perform inference processes conditioned on the input sample.

This paper aims to reduce the spatial redundancy of input images for classification. Our motivation lies in that there often exists considerable class-irrelevant region in the input image for general classification, especially those datasets with high-resolution images, \emph{e.g.} ImageNet~\cite{deng2009imagenet}. In fact, CNN can make the correct prediction according to a few class-discriminative patches, the total area of which is often much smaller than the original image. Ideally, if we can dynamically pinpoint those class-relevant patches and use them for efficient inference, the computation can be reduced significantly without the loss of accuracy. Thus the main challenge is how to pinpoint those important patches in time efficiently.

\begin{figure}[tbp]  
	\centering  
	\includegraphics[width=1\linewidth]{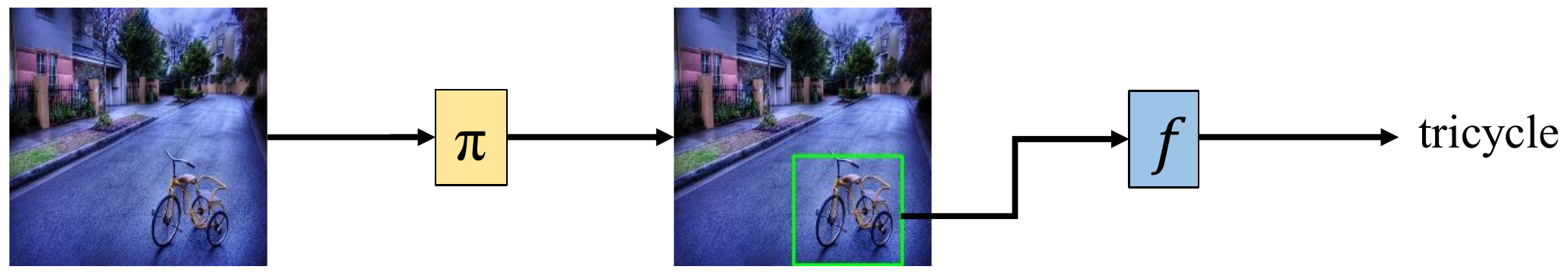}
	\vspace{-0.1cm}
	\caption{The toy example of the inference pipeline for accelerating image classification. For input images, the lightweight AnchorNet $\pi$ first captures class-discriminative  patches conditioned on the object properties. Then these localized patches with much smaller regions are further fed to a general classification network $f$ for class prediction.}  
	\label{pipeline}
	\vspace{-0.2cm}
\end{figure}
We propose a novel patch proposal network called AnchorNet that localizes semantic patches for downstream classification. The overall pipeline is shown in Fig.~\ref{pipeline}. To meet the requirements for accelerating inference, the upstream AnchorNet is equipped with two properties: (1) it is an extremely lightweight network that quickly captures patches with cheap computational costs; (2) localized patches should reflect class-discriminative features for potentially contributing to the final prediction. For implementing (2), we leverage a classic but widely ignored \textbf{\emph{Receptive Field}} (RF) mapping mechanism \emph{without intermediate paddings}. It can exactly map each spatial location from a high-level feature map to the corresponding patch in the input image, as shown in Fig.~\ref{map}. By deriving the class activation map (CAM)~\cite{zhou2016learning}, we can exactly evaluate each patch's contribution according to the corresponding activation value in CAM. In contrast, universal methods often adopt vanilla mapping using the relative proportion of the spatial size between the input image and its high-level feature map. This ignores that each high-level spatial location aggregates a larger region than its mapped region due to the intermediate paddings.

AnchorNet can be compatible with any classification network for dynamic inference. We conduct a sequential inference process, where at each step, the network receives a small patch localized by AnchorNet and makes the softmax-based prediction. If the confidence (\emph{i.e.} the maximum probability) exceeds the pre-defined threshold, the inference would be terminated and output the final prediction at the current step. Moreover, inspired by GFNet~\cite{wang2020glance}, we find a moderate portion of images could be correctly classified by simply resizing in practice. Therefore we introduce this cheap operation as the first step for inference before patch localization. In summary, the overall pipeline proceeds progressively with iteratively localizing patches by AnchorNet and then processing patches by the downstream network. The sequential inference process can be dynamically exited conditioned on the input patch. In practice, another advantage of our method lies in that it can readily adapt various trade-offs between accuracy and inference cost by simply tuning the confidence threshold.

We evaluate our AnchorNet by combining it with several widely applied networks on the highly competitive ImageNet dataset~\cite{deng2009imagenet}. Under the standard dynamic inference settings~\cite{huang2017multi}, combining our AnchorNet with ResNets~\cite{he2016deep} and DenseNets~\cite{huang2017densely} achieves better efficiency than SOTA MSDNet~\cite{huang2017multi}, RANet~\cite{yang2020resolution} and GFNet~\cite{wang2020glance}. 

\section{Related Works}
Most existing works aim to perform dynamic inference conditioned on the input image to reduce model redundancy, such as adaptively skipping layers or blocks~\cite{veit2018convolutional,wang2018skipnet}, dynamically selecting channels~\cite{hudrnet} and early exiting easy samples from intermediate auxiliary classifiers~\cite{teerapittayanon2016branchynet,huang2017multi}. Unlike the above works, our AnchorNet aims to dynamically reduce the spatial redundancy of input images without changing the architecture of the downstream classification network. Recently, RANet~\cite{yang2020resolution} considers reducing the spatial redundancy by adaptively resizing the input image according to its classification difficulty. GFNet~\cite{wang2020glance} adopts reinforcement learning to optimize a patch proposal network to accelerate image recognition. Some recent works~\cite{chen2021chasing,cordonnier2021differentiable} propose differentiable patch selection strategies. By contrast, our AnchorNet is a simple CNN-based feature extractor and is easy to be trained by the conventional cross-entropy loss. Based on CAM-based heatmaps~\cite{zhou2016learning}, our method can readily pinpoint meaningful patches to achieve more flexible dynamic inference without back-propagation. 

\section{Method}
\begin{figure}[tbp]  
	\centering  
	\includegraphics[width=1\linewidth]{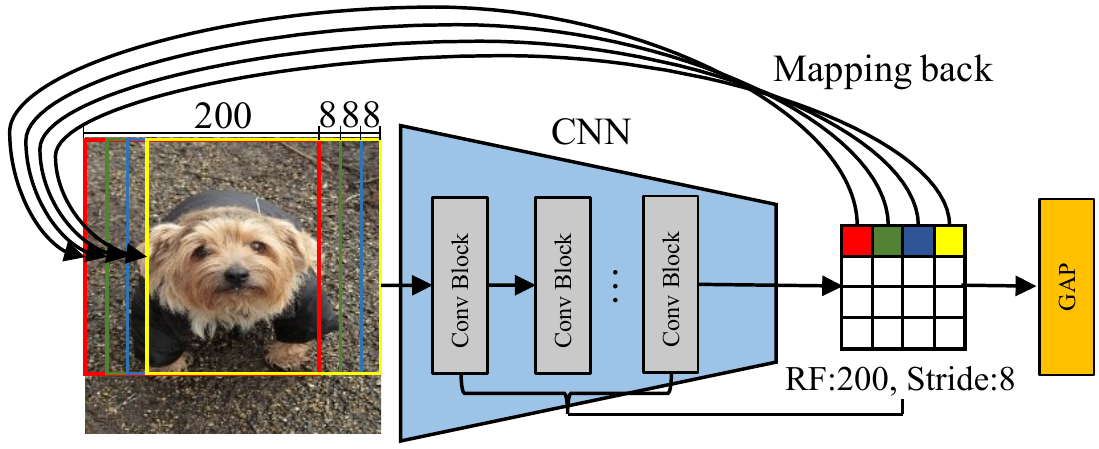}
	\caption{Example of exact patches mapping. We only depict the spatial dimension while omitting channels to better understand the spatial mapping rule. Best viewed in color.}  
	\label{map}
	\vspace{-0.3cm}
\end{figure}
\subsection{Review of Feature Alignment and RF Mapping}
Preliminarily, we review the rule of feature alignment from the perspective of Receptive Field (RF) mapping. Modern CNNs gradually decrease the spatial resolution of the input image by several convolutional blocks until the global average pooling (GAP) layer. Kernel size and stride are critical hyperparameters in convolutional layers for constructing the accumulated RF. We also remark that intermediate paddings over feature maps would destroy the exact RF mapping rule. That is to say, the CNN without paddings can provide exact spatial mappings from the high-level feature map (before GAP) to the input image, which we tag as an \emph{\textbf{interpretable}} network in this paper. As illustrated in Fig.~\ref{map}, assuming that an interpretable CNN receives an input image with $H\times W$ resolutions and
has accumulated $K\times K$ RF size and $S$ strides before GAP, we will
obtain $[\left \lfloor (H-K)/S \right \rfloor+1] \times [\left \lfloor (W-K)/S \right \rfloor+1]$ high-level spatial locations. Here, $H$ and $W$ denote the height and width, respectively. Each location can be mapped back to a patch region with the size of $K\times K$ in the original input image. For example, given an input image with $224\times 224$ pixels, a CNN with accumulated $200\times 200$ RF size and $8$ strides before GAP can generate a feature map with $4\times4=16$ spatial locations.

We further review the calculation of accumulated RF. For simplicity, we suppose the height and width of kernel size and stride are equal. We denote the accumulated RF and stride from the first to the $(l-1)$-th convolutional layer as $K_{l-1}$ and $S_{l-1}$, respectively. The kernel size and stride of the current $l$-th convolutional layer are $k_{l}$ and $s_{l}$, respectively. The accumulated RF $K_{l}$ and stride $S_{l}$ until the current $l$-th layer are formulated as:
\begin{align}
	&K_{l}=K_{l-1}+(k_{l}-1)*S_{l-1},\ K_{1}=k_{1} \notag \\
	&S_{l}=S_{l-1}*s_{l},\ S_{1}=s_{1} \notag \\
	&s.t.\ K_{l} \leqslant \min(H, W) 
\end{align} 
Due to the no-padding design, the spatial mapping from a high-level feature map to the original input is guaranteed to be exact.
\begin{figure*}[tbp]  
	\centering  
	\includegraphics[width=1.\linewidth]{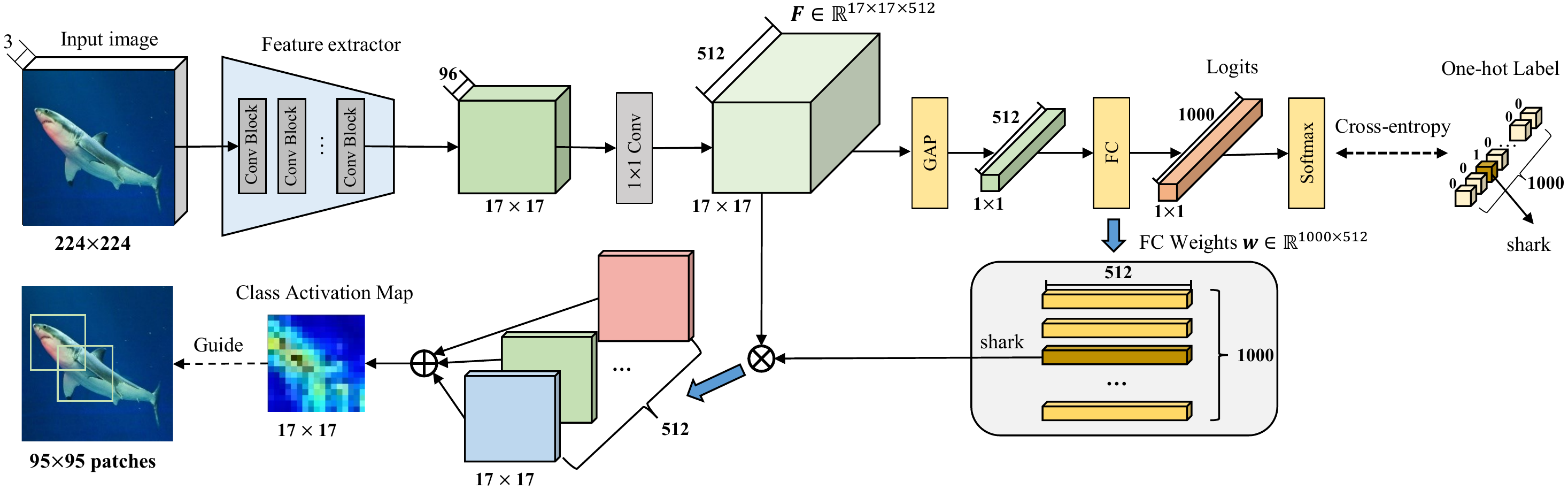}
	\caption{The overview of our proposed AnchorNet. Here, the boxes represent various feature maps, where the spatial dimension is tagged below. We derive class activation maps~\cite{zhou2016learning} from AnchorNet to guide the patch selection. }  
	\vspace{-0.2cm}
	\label{anchornet}
\end{figure*}
\begin{table}[t]
	\centering
	\caption{The architectural settings of the extractor. 'OR' denotes the output resolution, 'Exp' denotes the expansion ratio, 'Out' denotes the output channels,  '$s$' denotes the stride of the current convolution, and 'RF' represents the accumulated RF size until the current layer. The detailed architectures with hyper-parameter explanations are shown in EfficientNet~\cite{Mingxing2019EfficientNet}.}
	
	\begin{tabular}{c|c|c|c|c|c}  
		\hline
		OR & Operator, kernel & Exp &Out &$s$&RF \\
		\hline
		$111^{2}$       & Conv,3$\times$3    &-&16&2&$3^{2}$   \\
		$55^{2}$      & MBConv,3$\times$3& 1&16&2&$7^{2}$   \\
		$27^{2}$      & MBConv,3$\times$3& 3 &24&2&$15^{2}$   \\
		$25^{2}$      & MBConv,3$\times$3& 4 &24&1&$31^{2}$   \\
		$23^{2}$      & MBConv,3$\times$3& 4 &48&1&$47^{2}$   \\
		$21^{2}$      & MBConv,3$\times$3& 2 &96&1&$63^{2}$   \\
		$19^{2}$      & MBConv,3$\times$3& 1.5 &96&1&$79^{2}$   \\
		$17^{2}$      & MBConv,3$\times$3& 1.5 &96&1&$95^{2}$   \\
		\hline
	\end{tabular}
	\vspace{-0.2cm}
	\label{details}
\end{table}

\subsection{AnchorNet}
We develop a lightweight patch proposal network called AnchorNet. It can provide the interpretable patch-wise contribution while adaptively
localizing informative patches conditioned on the input image. Compared with the design thought of popular networks, the differences are the properties of carefully designed RF without paddings and exact patch mappings.

To pursue high efficiency, we utilize the \emph{MBConv} module proposed in EfficientNet~\cite{Mingxing2019EfficientNet} as the meta-architecture to stack a feature extractor, where we set padding $0$ across all convolutional layers. The feature extractor's overall architecture and configuration details are shown in Fig.~\ref{anchornet} and Table~\ref{details}, respectively. Then we attach a $1\times 1$ convolution for expanding feature channels and a linear softmax classifier to learn task-aware information supervised by one-hot ground-truth labels. Inspired by Wang \emph{et al.}~\cite{wang2020glance}, we construct the accumulated RF of AnchorNet as $95\times 95$ for ImageNet~\cite{deng2009imagenet} classification. It means that AnchorNet can perform spatial pixel-wise mappings with a patch size of  $95\times 95$ guided by the high-level CAM~\cite{zhou2016learning}. Since the accumulated stride is $2^{3}=8$, AnchorNet can map the original $224\times 224$ image to $17^{2}=289$ candidate patches. In terms of complexity, FLOPs\footnote{The indicator of FLOPs denotes the number of floating-point operations.} of the resulting AnchorNet is only 60 MFLOPs, which is about 1.5\% of the 4089 MFLOPs of the widely used ResNet-50.  
\subsection{Localizing Semantic Patches}
\label{Localizing}
As shown in Fig.~\ref{anchornet}, we utilize Class Activation Map (CAM)~\cite{zhou2016learning} to evaluate the importance of class-aware spatial locations to the final predictions. We denote the high-level feature map before GAP layer as $\mathbf{F}\in \mathbb{R}^{H\times W\times C}$ with $C$ channels, where $\mathbf{F}_{c}\in \mathbb{R}^{H\times W}$ is the $c$-th spatial map along the channel dimension. The weight tensor of the linear classifier is $\mathbf{w}\in \mathbb{R}^{N\times C}$, where $N$ is the number of classes, $w_{n,c}$ denotes the value from the $n$-th row and $c$-th column of weight matrix $\mathbf{w}$. The CAM of the $n$-th class can be formulated as:
\begin{equation}
	M_{n}=\sum_{c=1}^{C}w_{n,c}\mathbf{F}_{c}
\end{equation}
Each activation value of the spatial location over the CAM $M_{n}\in \mathbb{R}^{H\times W}$ can represent the importance of the mapped image patch contributing to the class $n$. In practice, we may utilize multiple patches to fully cover the class discriminative features. However, simply localizing those patches with maximum activations may result in excessive overlaps. Therefore, we propose a simple yet effective heuristic patch selection method. It can ensure the localized patches that are not only important but also partly separated to cover more information. Specifically, we introduce Intersection-over-Union (IoU) to quantify the intersection between two given patches A and B, \emph{i.e.} $IoU(A,B)=|A\cap B|/|A \cup B|$, where $|\cdot|$ calculates the pixel number of the given region. We denote the localized patch collection as $S$ and visit each spatial location over $M_{n}$ from large to small according to the activation value. If IoU between the patch mapped from the currently visited location and any localized patch in $S$ is less than the threshold $\theta$, we put this patch into $S$. Finally, we can obtain $S$ as the resulting patch collection. This idea behind the patch selection method is inspired by the widespread Non-Maximum Suppression (NMS) for object detection. 

\begin{figure}[tbp]  
	\centering  
	\includegraphics[width=1.\linewidth]{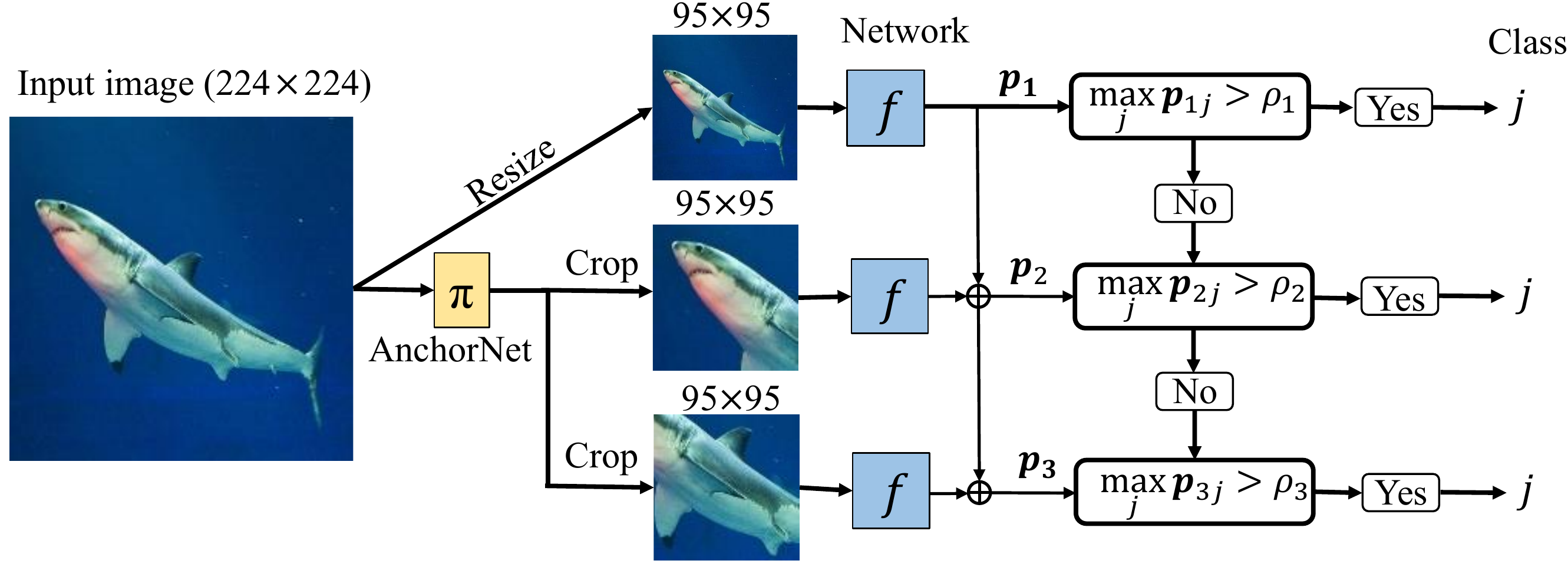}
	\caption{The overall framework of the sequential inference pipeline with the length of $T=3$.}  
	\label{Sequential}
	
\end{figure}
\begin{algorithm}[tb]
	\caption{Sequential Inference Process}
	\label{SIP}
	\textbf{Input}: a sequence of images $\mathcal{X}=\{\bar{\bm{x}}_{1},\bar{\bm{x}}_{2},\cdots,\bar{\bm{x}}_{T}\}$\\
	\textbf{Parameter}: network $f$, threshold $\rho_{1}\sim \rho_{T-1}$, length $T$ \\
	\textbf{Output}: the predicted class $j$
	
	\begin{algorithmic}[1] 
		\FOR{$t=1:T$}
		\STATE \rm{\textbf{if}} $t=1$ \rm{\textbf{then}} $\bm{p}_{t}=f(\bar{\bm{x}}_{t})$
		\STATE \rm{\textbf{else}} $\bm{p}_{t}=\bm{p}_{t-1}+f(\bar{\bm{x}}_{t})$
		\STATE \rm{\textbf{if}} $t<T$ and $\max_{j}{\bm{p}_{tj}}>\rho_{t}$ \rm{\textbf{then}} \rm{\textbf{return}} $j$
		\ENDFOR
		\STATE \textbf{return} $\arg \max_{j}{\bm{p}_{Tj}}$ 
	\end{algorithmic}
\end{algorithm}
\subsection{Sequential Inference Pipeline}
For easy notation, we denote AnchorNet as $\pi$ and the downstream classification network as $f$. Given an input image $\bm{x}$, the overall framework of sequential inference pipeline is illustrated in Fig.~\ref{Sequential}. It can be formulated as iteratively processing a sequence of images $\mathcal{X}=\{\bar{\bm{x}}_{1},\bar{\bm{x}}_{2},\cdots,\bar{\bm{x}}_{T}\}$. Yang \emph{et al.}~\cite{yang2020resolution} found that a moderate portion of images can be correctly classified by simply resizing. Thus we first resize an original $224\times 224$ images to a $95\times 95$ image denoted as $\bar{\bm{x}}_{1}$. The remaining image patches of $\{\bar{\bm{x}}_{2},\cdots,\bar{\bm{x}}_{T}\}$ are localized by AnchorNet. At the $t$-th step of dynamic inference, if the softmax confidence $\max_{j}{\bm{p}_{tj}}$ is larger than the pre-defined threshold $\rho_{t}$, the sequential inference would be terminated with a predicted class $j$. To make use of patch information, we formulate
$\bm{p}_{t}=f(\bar{\bm{x}}_{1})+f(\bar{\bm{x}}_{2})+\cdots+f(\bar{\bm{x}}_{t})$, $1\le t\le T$. The overall inference process is shown in Alg.~\ref{SIP}.

\subsection{Training Strategy}
We conduct a two-stage training scheme to train AnchorNet $\pi$ at \textbf{Stage I} and the downstream classification network $f$ at \textbf{Stage II}.

\textbf{Stage I}: Given an input image $\bm{x}$, AnchorNet $\pi$ outputs the softmax-based class probability distribution  $\pi(\bm{x})$. We utilize the standard cross-entropy loss $\mathcal{L}_{ce}$ with ground-truth labels to train $\pi$:
\begin{equation}
	\mathcal{L}_{\pi}=\frac{1}{|\mathcal{D}|}\sum_{(\bm{x},y)\in \mathcal{D}}\mathcal{L}_{ce}(\pi(\bm{x}),y),
\end{equation}
where $\mathcal{D}$ is the training set, $y$ is the ground-truth label.

\textbf{Stage II}: Given an input image $\bm{x}$ and the pre-trained AnchorNet $\pi$, we can naturally obtain the sequence of relevant patches $\mathcal{X}=\{\bar{\bm{x}}_{1},\bar{\bm{x}}_{2},\cdots,\bar{\bm{x}}_{T}\}$. We initialize the network $f$ with pre-trained weights over the original $224\times 224$ images. We then finetune $f$ with these patches in $\mathcal{X}$ by standard $\mathcal{L}_{ce}$:
\begin{equation}
	\mathcal{L}_{f}=\frac{1}{|\mathcal{D}|}\frac{1}{|\mathcal{X}|}\sum_{(\bm{x},y)\in \mathcal{D}}\sum_{\hat{\bm{x}}\in \{\bar{\bm{x}}_{1}\}}\mathcal{L}_{ce}(f(\hat{\bm{x}}),y),
	\label{global}
\end{equation}
\begin{equation}
	\mathcal{L}_{f}=\frac{1}{|\mathcal{D}|}\frac{1}{|\mathcal{X}|}\sum_{(\bm{x},y)\in \mathcal{D}}\sum_{\hat{\bm{x}}\in \{\bar{\bm{x}}_{2},\cdots,\bar{\bm{x}}_{T}\}}\mathcal{L}_{ce}(f(\hat{\bm{x}}),y),
	\label{local}
\end{equation}
where $f(\hat{\bm{x}})$ is the predicted class probability distribution from image $\hat{\bm{x}}$. Following Wang \emph{et al.}~\cite{wang2020glance}, we finetune two networks of $f$ using Eq.~(\ref{global}) and Eq.~(\ref{local}) to classify the resized image $\bar{\bm{x}}_{1}$ and localized patches $\{\bar{\bm{x}}_{2},\cdots,\bar{\bm{x}}_{T}\}$, respectively.

\section{Experiments}
\subsection{Experimental Settings}
\textbf{Dataset and training setup.} We conduct evaluations of our dynamic inference pipeline on ImageNet~\cite{deng2009imagenet} benchmark, a large-scale dataset containing 1.2 million training images and 50K  validation images in 1000 classes. We adopt the standard data augmentation schemes~\cite{he2016deep}, and the resolution of each input image is $224\times 224$.  The SGD optimizer trains all networks with a momentum of 0.9, a weight decay of $1\times 10^{-4}$ and a batch size of 256. The initial learning rate is 0.1 and is divided by 10 at the 30-th, 60-th and 90-th epoch within the total 100 epochs for AnchorNet training. We finetune the downstream network $f$ using a cosine learning rate from 0.01 within the total 30 epochs. The input size of $f$ is $95\times 95$.

\textbf{Dynamic inference setup.} Following~\cite{huang2017multi}, we conduct two patterns for dynamic inference:  (1) \emph{Budgeted Batch Classification}, which requires the model to finish the predictions of a set of test samples within a specific computational budget. (2) \emph{Anytime prediction}, which requires the model to predict each test sample with a finite computational budget. In this paper, for (1), we tune the confidence thresholds $\rho_{1}\sim \rho_{T-1}$ to finish adaptive inference for various test samples to satisfy the  computational budget; for (2), we force all test samples to finish predictions from the same exit point.

\textbf{Hyper-parameters setup.} We set the maximum patch number as $T=5$ so that the total area of all $95\times 95$ patches is smaller than that of the original $224\times 224$ image. We empirically set the IoU threshold $\theta=0.3$ to localize patches in Section~\ref{Localizing}. 

\begin{figure}[tbp]  
	\centering  
	\includegraphics[width=1.\linewidth]{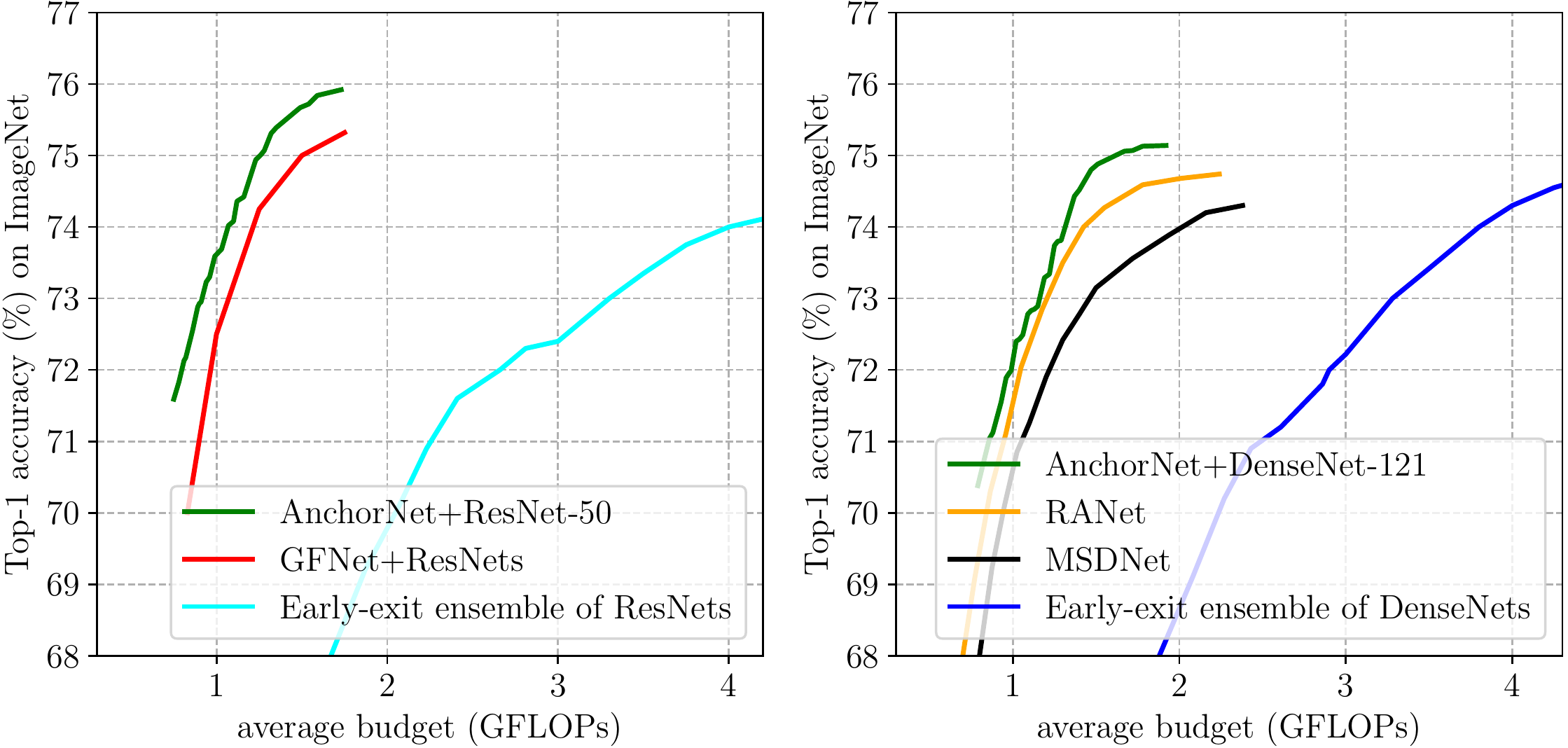}
	\caption{Comparison of trade-offs between accuracy and computational budget based on ResNets (\emph{left}) and DenseNets (\emph{right}) under \emph{budgeted batch classification}.} 
	\label{Budgeted}
\end{figure}

\subsection{Comparison with Dynamic Inference Methods}
\begin{figure}[tbp]  
	\centering  
	\includegraphics[width=1.\linewidth]{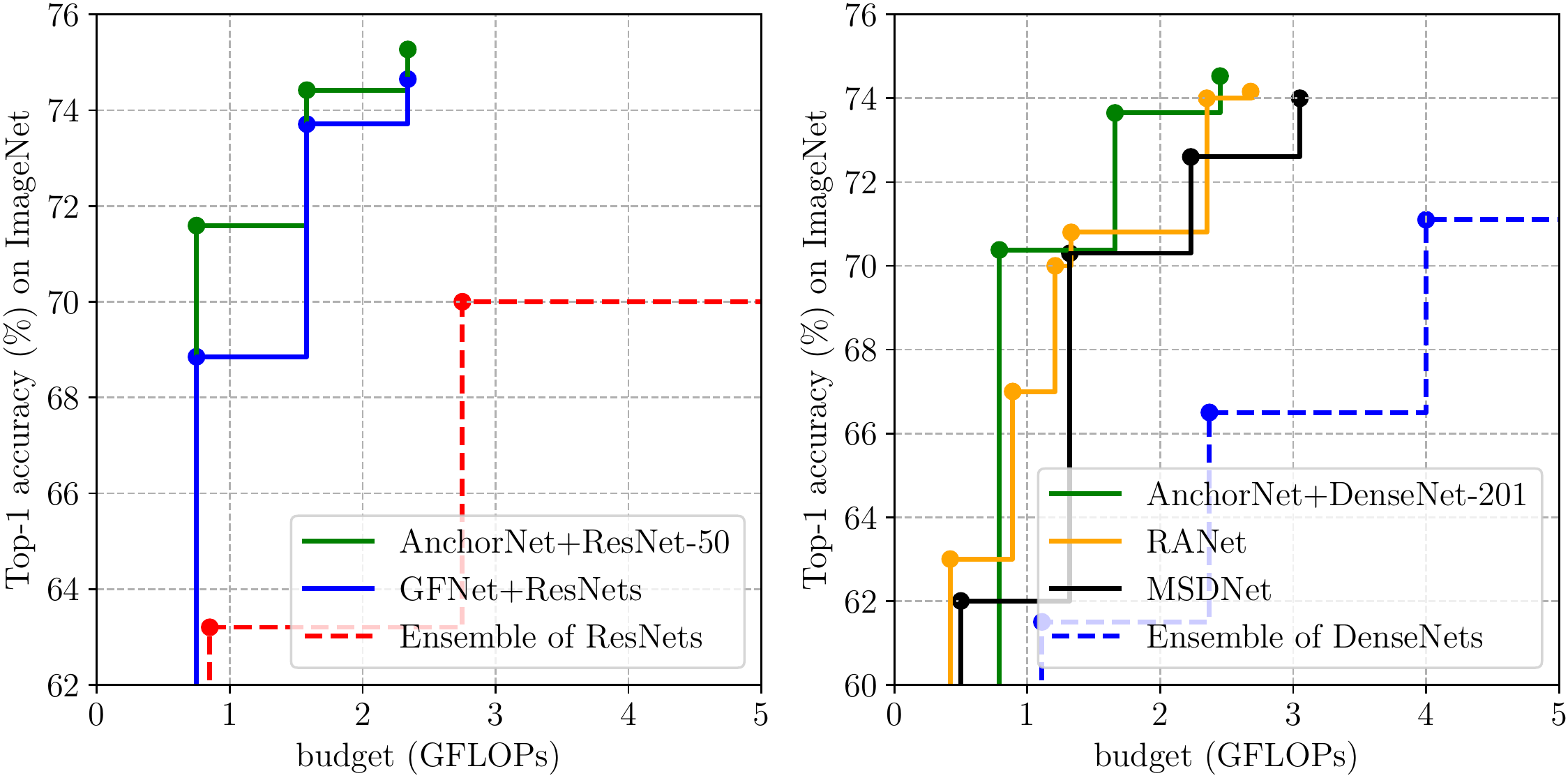}
	\caption{Comparison of trade-offs between accuracy and computational budget based on ResNets (\emph{left}) and DenseNets (\emph{right}) under \emph{anytime prediction}. } 
	\label{dynamic}
	\vspace{-0.3cm}
\end{figure}

We conduct a comprehensive comparison with state-of-the-art dynamic inference methods, such as MSDNet~\cite{huang2017multi}, RANet~\cite{yang2020resolution} and GFNet~\cite{wang2020glance}, under fair dynamic inference settings. We use widely applied ResNets~\cite{he2016deep} and DenseNets~\cite{huang2017densely} as downstream classification networks. Moreover, we also evaluate two highly competitive baselines, including ensembles of ResNets and ensembles of DenseNets with various depths~\cite{huang2017multi}. The comparisons of trade-offs between accuracy and computational budget are shown in Fig.~\ref{dynamic}. In general, all dynamic inference methods can significantly outperform other baseline ResNets and DenseNets networks.

\textbf{Budgeted batch classification}. As shown in Fig.~\ref{Budgeted}, our AnchorNet framework consistently achieves better accuracies than recent dynamic inference methods under the same computational budgets. With the range of $0.75\sim 1.7$ GFLOPs for ResNet, our AnchorNet achieves $71.6\%\sim 75.9\%$ top-1 accuracies, outperforming GFNet by $0.5\%\sim 2\%$ gains. Results on DenseNets show a similar observation that AnchorNet surpasses MSDNet and RANet consistently.

\textbf{Anytime prediction}.  As shown in Fig.~\ref{dynamic}, the results under anytime prediction also demonstrate the superiority of our AnchorNet. It obtains $2.7\%$, $0.7\%$ and $0.6\%$ improvements compared to GFNet over the same three exit points. Extended 
\begin{figure}[tbp]  
	\centering  
	\includegraphics[width=1.\linewidth]{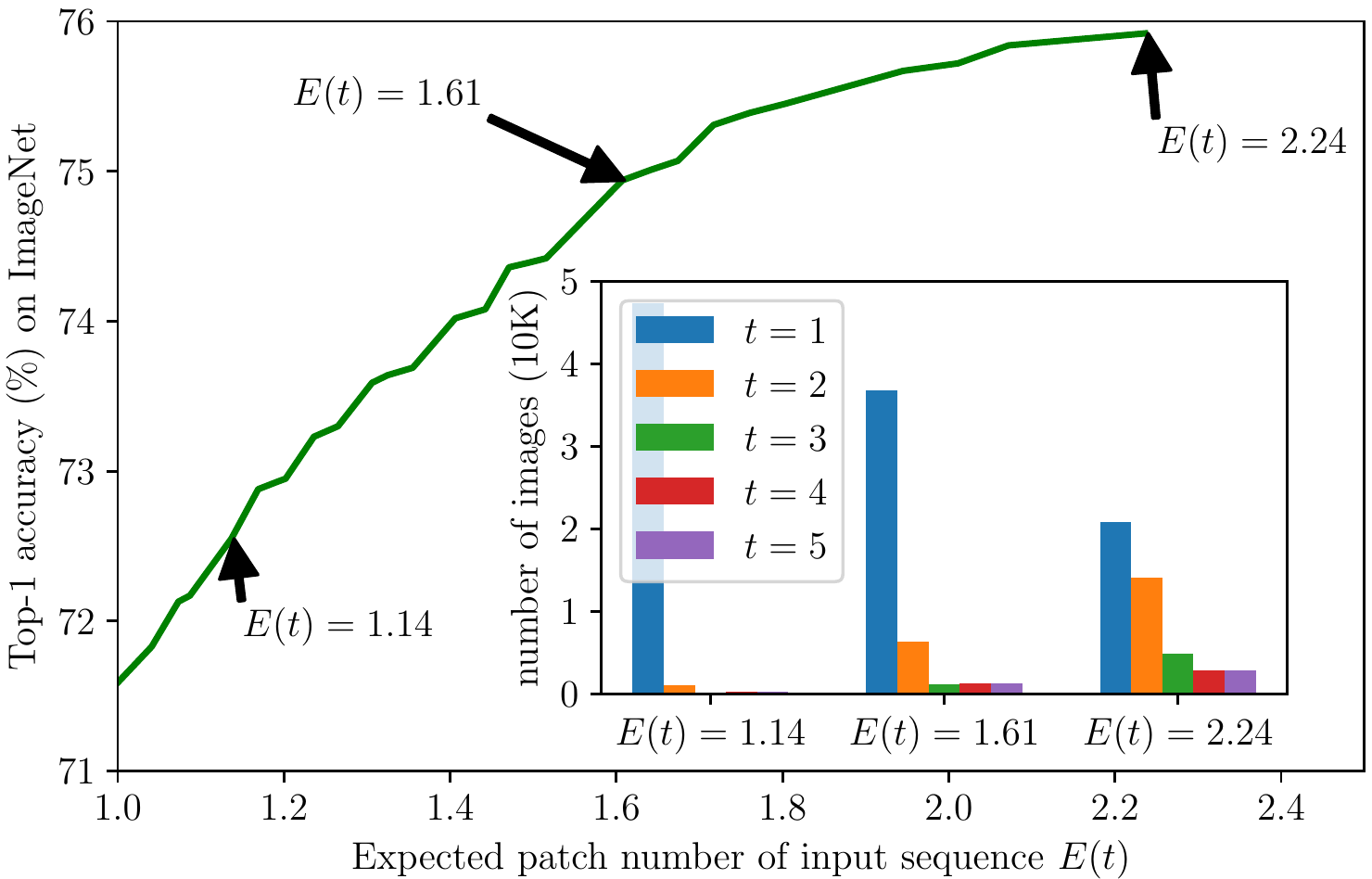}
	\caption{Top-1 accuracy v.s. the expected patch number of input sequence $E(t)$ under \emph{budgeted batch classification}.} 
	\vspace{-0.1cm}
	\label{expected_patch}
\end{figure}
\begin{table}[t]
	\centering
	\caption{Average inference time with a batch size of 1K.}
	\vspace{-0.2cm}
	\label{time}
	\resizebox{0.43\textwidth}{!}{
		\begin{tabular}{c|ccc|c}  
			\hline
			Method&\multicolumn{3}{c|}{+AnchorNet (Ours)}  & DenseNet-201 \\
			\hline
			Time (ms/img)&0.6&1.1&1.6&1.7\\
			FLOPs (G)&0.8&1.6&2.4&4.3\\
			Accuracy (\%)&70.38&73.65&74.53& 76.93\\
			\hline
	\end{tabular}}
	\vspace{-0.9cm}
\end{table}
\begin{table}[H]
	\centering
	\caption{Accuracy against random patches on ResNet-50.}
	\vspace{-0.2cm}
	\resizebox{0.43\textwidth}{!}{
		\begin{tabular}{lcccccc}  
			\hline
			Patch number&1&2&3&4&5\\
			\hline
			Random& 58.50& 66.34 &68.55& 70.30&70.81\\
			AnchorNet& \textbf{71.59} & \textbf{74.42} & \textbf{75.27}& \textbf{75.48} & \textbf{75.65} \\
			\hline
	\end{tabular}}
	\label{random}
	\vspace{-0.2cm}
\end{table}
\noindent experiments on DenseNet show that AnchorNet achieves similar accuracies with RANet and MSDNet using less computational budgets. The results demonstrate that AnchorNet is a more effective dynamic inference pattern to pursue higher computational efficiency.

\subsection{Ablation Study and Discussion}
\label{section_ablation}
\textbf{Actual inference time.} In Table~\ref{time}, we test the actual inference time on DenseNet-201 using a single NVIDIA 3090 GPU under \emph{anytime prediction}. We
find that the ratio of the actual acceleration on time is
lower than that of the theoretical acceleration on FLOPs due to the implementation of PyTorch library. We hope the acceleration gap could be reduced benefiting from the development of deep learning frameworks. 

\textbf{Comparison with random patches.} We compare AnchorNet with the simple random policy where all patches are uniformly chosen from the image. In Table~\ref{random}, AnchorNet outperforms the random policy across various budgets, indicating the former captures semantic object regions than the latter.

\textbf{Impact of patch number.} In Fig.~\ref{expected_patch}, we observe more significant accuracy gains as the expected patch number of input sequence $E(t)$ increases on ResNet-50. We further sample several points and show their distributions of the numbers of images exited from $t=1\sim 5$ stages.  By improving thresholds $\rho_{1}\sim \rho_{4}$, more samples tend to exit from later stages with higher softmax confidences, leading to higher accuracy.

\begin{figure}[tbp]  
	\centering  
	\includegraphics[width=1.\linewidth]{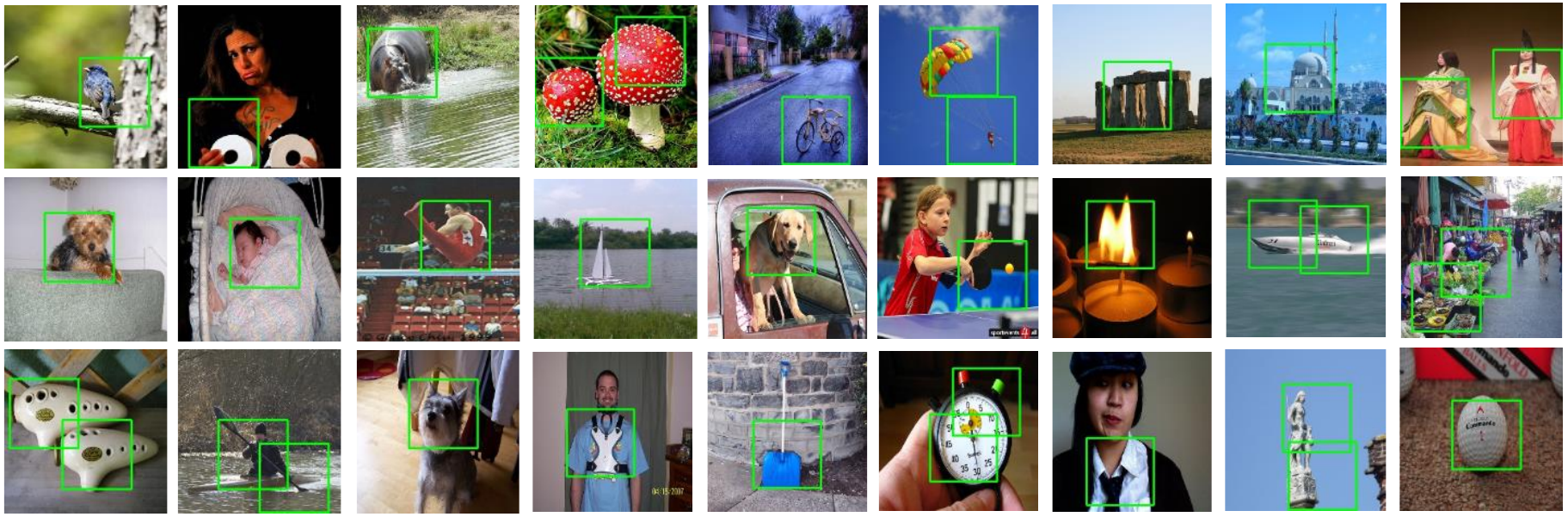}
	\vspace{-0.4cm}
	\caption{Visualization of localized patches on ImageNet.} 
	\label{patches}
	\vspace{-0.2cm}
\end{figure}

\textbf{Visualization of localized patches.} As shown in Fig.~\ref{patches}, we can observe that AnchorNet can adaptively cover class-discriminative features conditioned on the input image. This further verifies the rationality of using semantic patches for downstream classification.

\section{Conclusion}
We propose a dynamic inference framework for accelerating image classification by introducing an upstream lightweight patch proposal network called AnchorNet. The interpretable AnchorNet can efficiently localize those class-discriminative patches in the input image to reduce spatial redundancy.   We then sequentially feed these patches into a downstream general classification network, which only requires processing a much smaller region than the original input. Our AnchorNet can be compatible with any classification network and consistently improve computational efficiency over widely used networks in practice.

\bibliographystyle{IEEEbib}
\bibliography{icme2022template}

\end{document}